\documentclass[journal]{IEEEtran}
\usepackage[font=small,skip=1pt]{caption}
\IEEEoverridecommandlockouts
\usepackage{textcomp}

\usepackage{amsmath} 
\usepackage{amssymb} 
\usepackage{amsfonts} 
\usepackage{graphicx} 
\usepackage{multirow} 
\usepackage{booktabs} 
\usepackage{algorithm} 
\usepackage{algpseudocode} 
\usepackage{cite} 
\usepackage{xcolor} 
\usepackage{float} 
\usepackage{hyperref} 
\usepackage{amsthm} 
\usepackage{bm} 
\usepackage{enumitem} 
\usepackage{inputenc} 
\usepackage[english]{babel} 
\usepackage{lineno} 
\usepackage{mathdots} 
\usepackage{tabularx} 
\usepackage{threeparttable} 
\usepackage{epstopdf} 
\usepackage{fontenc} 
\usepackage{pifont} 
\usepackage{srcltx} 
\usepackage{subfig} 

\begin{document}
\title{Heuristical Comparison of Vision Transformers Against Convolutional Neural Networks for Semantic Segmentation on Remote Sensing Imagery}
\author{Ashim Dahal, Saydul Akbar Murad, and Nick Rahimi, \IEEEmembership{Member, IEEE}

\thanks{ A. Dahal, S. A. Murad, and N. Rahimi Authors are with the School of Computing Sciences and Computer Engineering, University of Southern Mississippi, Hattiesburg, USA (ashim.dahal@usm.edu, saydulakbar.murad@usm.edu, nick.rahimi@usm.edu).}
}

\maketitle

\begin{abstract}
Vision Transformers (ViT) have recently brought a new wave of research in the field of computer vision. These models have done particularly well in the field of image classification and segmentation. Research on semantic and instance segmentation has emerged to accelerate with the inception of the new architecture, with over 80\% of the top 20 benchmarks for the iSAID dataset being either based on the ViT architecture or the attention mechanism behind its success. This paper focuses on the heuristic comparison of three key factors of using (or not using) ViT for semantic segmentation of remote sensing aerial images on the iSAID. The experimental results observed during the course of the research were under the scrutinization of the following objectives: 1. Use of weighted fused loss function for the maximum mean Intersection over Union (mIoU) score, Dice score, and minimization or conservation of entropy or class representation, 2. Comparison of transfer learning on Meta’s MaskFormer, a ViT-based semantic segmentation model, against generic UNet Convolutional Neural networks (CNNs) judged over mIoU, Dice scores, training efficiency, and inference time, and 3. What do we lose for what we gain? i.e., the comparison of the two models against current state-of-art segmentation models. We show the use of the novel combined weighted loss function significantly boosts the CNN model’s performance capacities as compared to transfer learning the ViT. The code for this implementation can be found on \href{https://github.com/ashimdahal/ViT-vs-CNN-Image-Segmentation}{https://github.com/ashimdahal/ViT-vs-CNN-Image-Segmentation}.
\end{abstract}

\begin{figure}[!ht]
    \centering
    \includegraphics[width=1\linewidth]{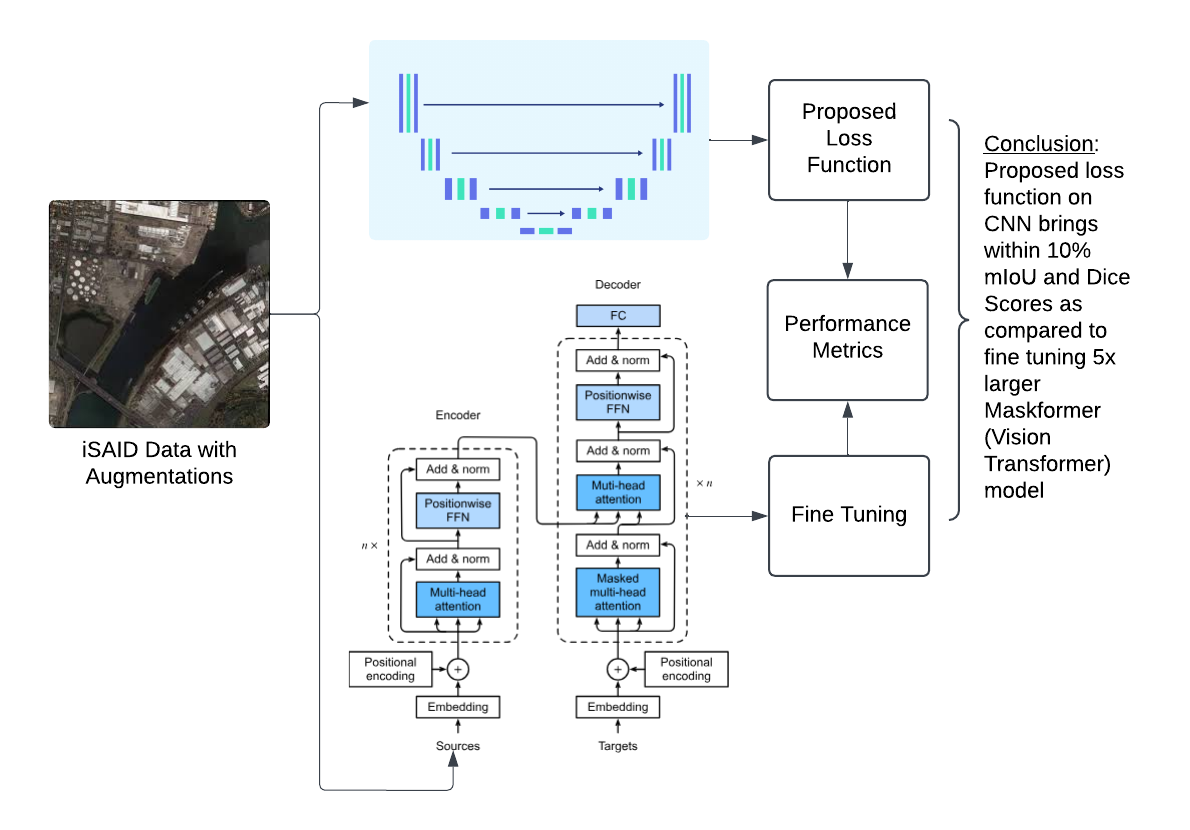}
\end{figure}

\begin{IEEEkeywords}
Vision Transformers, Semantic Segmentation, Convolutional Neural Networks (CNNs), MaskFormer and Remote Sensing
\end{IEEEkeywords}

\section{Introduction}
The introduction of Transformers \cite{transformer} has changed the research landscape when it comes to attention mechanisms on Natural Language Processing (NLP) tasks. However, this potential wasn't fully capitalized in computer vision tasks until recently when Dosovitskiy et al. implemented the attention mechanism of transformers in their seminal paper \cite{ViT} ever since ViT has been one of the fundamental areas of research in computer vision. Although initially proposed for image classification tasks, like CNNs in their inception days, they soon turned out to be one of the best-performing architectures for image segmentation tasks as well.

Image segmentation refers to the task of classifying each pixel of an image into a category. It can be said that image segmentation is a type of classification as well, but the role, approach and method of each subjects varies in themselves. In specific to the scope of this paper, semantic segmentation means to group certain objects in an image to a class among the given n classes in the dataset. Instance segmentation, similarly, would be to segment each object present in an image as a distinct item in itself.

As with any deep learning computer vision tasks, image segmentations were best done by models that were capable of capturing, encoding, and decoding essential patterns of the input image, mainly the implementation of UNet style architectures in CNNs \cite{TreeUnet,UnetFirst,segnet,multiresunet,unet_orig,context_encoding}. Specific to the scope of this paper, in the past few years researchers have utilized the UNet CNNs in the iSAID dataset\cite{waqas2019isaid}, which is built on top of the DOTA\cite{dota} dataset, to do semantic segmentations  \cite{isaid_unet1,isaid_unet2,isaid_unet3,isaid_unet4}. One of the main pitfalls of such datasets is the background class\cite{pitfall}. If not handled properly during the training process, there is a good chance to see high evaluation metrics on mIoU since the model would easily overfit on the most common class, which is the unlabelled class in this case.

Although the traditional deep learning technique of employing a UNet-based Convolutional Neural Network (CNN) remains the cornerstone of many segmentation tasks, recent trends in computer vision research indicate a significant shift towards Transformer-based architectures, particularly the Vision Transformer (ViT) and its variants. This shift is driven by the inherent ability of Transformer models to capture long-range dependencies through self-attention mechanisms, a feature that CNNs typically struggle with due to their localized receptive fields. The increasing preference for ViT models is exemplified by the fact that among the top 20 benchmark models for the iSAID dataset as listed on Papers with Code\cite{paper_with_code}, the top five employ either ViT, attention-based CNNs, or a hybrid combination of both \cite{isaid_vit1,isaid_vit2,isaid_vit3,isaid_vit4}. These models leverage the powerful self-attention mechanism to refine segmentation masks by focusing on relevant image regions.

In this paper, we introduce a novel loss function that integrates maximizing mean Intersection over Union (mIoU) and Dice score, while preserving entropy to ensure robust mask predictions. This new loss function is integrated into the UNet framework to improve its ability to model complex spatial relationships in images. This unique formulation allows for better generalization to unseen data by maintaining a balance between maximizing overlap with the ground truth masks and preventing over-segmentation. In addition, we investigate various data augmentation techniques since the number of samples is relatively lower in the iSAID\cite{waqas2019isaid} dataset. In parallel, we provide a direct comparison between training a UNet CNN model from scratch and fine-tuning Meta AI's widely adopted MaskFormer\cite{maskformer}, a ViT-based model that has achieved state-of-the-art results in semantic segmentation tasks; opensourced in Hugging Face\cite{hf}. 

This paper focuses on three key objectives throughout its experimentation and analysis:
\begin{itemize}
    \item Propose a combined weighted loss function to maximize mIoU and Dice while preserving entropy


    \item Analyze the impact on efficiency during training and inference in both architectures

    \item Benchmark and test inference capabilities of both models against current state-of-the-art iSAID benchmarks on unseen data
\end{itemize}

The rest of the paper is laid out in the following order, section \ref{lit_review} contains the current state of art and previous work on the field. Section \ref{methodology} discusses our approach to the given objectives, then \ref{results} effectively binds the results with the research question. Section \ref{conclusion} would then conclude the paper with some ending thoughts and future direction for the research field.

\section{Literature Review}\label{lit_review}
%
%
%
%
%
\begin{figure*}[!ht]
    \centering
    \includegraphics[width=\textwidth]{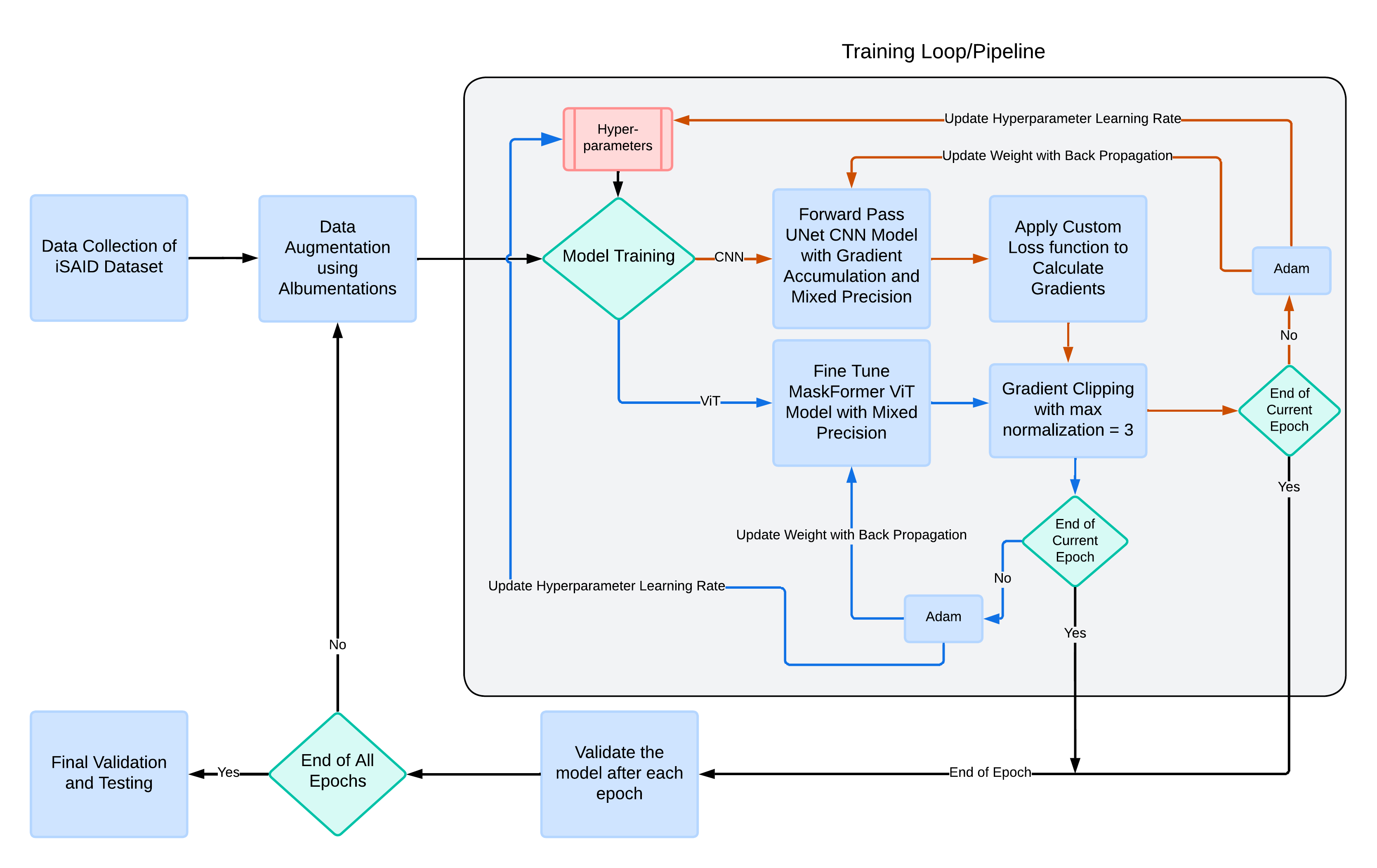}
    \captionsetup{justification=centering}
    \caption{Brief Overview of the Training and Validation Lifecycle}
    \label{fig:methodology}
\end{figure*}

Noh et al. \cite{UnetFirst} first proposed the UNet style model by training a deep deconvolutional layer on top of the VGG-16\cite{vgg16} CNN. Their work provided a path for later iterations of UNet-based systems on the iSAID dataset. Regmi \cite{isaid_unet1} presented their unsupervised model FreeSOLO for the iSAID dataset as a state-of-the-art. They deploy unsupervised learning, such as segmentation on the iSAID and other remote imagery data. However, the author fails to recognize the disparity between the claims and the results. The model fails to acknowledge small items on the dataset. This mainly comes from the author's choice of image preprocessing which is to bilinearly transform the image into (3, 256, 256) pixel inputs. This is not viable for the iSAID dataset because the pictures in the datasets range from size (3, 800, 800) to (3, 4000, 13000) and downsampling in such cases would result in the loss of a lot of necessary information and pattern that the model would need to make a robust prediction on all images. The results presented by the author also present some doubts since the ${AP_{50}}$ percentage is not scaled to the standard value, and a score of 0.9\% to 3.5\% would rather indicate a poor model by the accepted standards. Also the Dice score and the IoU score are only presented for the backbone models, which too are below 65\%, which is within range of the current state of the art ViT models and what our UNet CNN and MaskFormer models have surpassed.

In \cite{isaid_unet3}, Wang et al. show the best results obtained by the CNN models and ViT models after fine-tuning on seminal work by Xiao et al.; the UperNet model\cite{uppernet}. Both of their top results, on UNet and ViT, in the iSAID dataset come from fine-tuning the UperNet model. The best score for the CNN approach comes from fine-tuning the UperNet model pre-trained on the imagenet\cite{imagenet} dataset using a ResNet-50\cite{resnet} backbone whereas their best IoU score for ViT, which is also the overall best, comes form fine-tuning the UperNet model on the same dataset with the ViTAEv2-S\cite{zhang2023vitaev2} backbone; the scores were 62.54 and 66.26 respectively. While it is an impressive feat to have achieved such a high average IoU score in the iSAID dataset, we believe and show that the results could be improved for a much higher mIoU score. The authors also don't mention the Dice scores for their experiments. While the Dice score in most cases is very similar, or within the 5\% range, to the mIoU, having a fixed number would have been better for direct comparison with other approaches.

After the introduction of ViT\cite{ViT}, papers like \cite{isaid_vit4}\cite{isaid_vit_imp} have emerged in the iSAID dataset, which focuses on using the ViT in their approach. Papers like \cite{isaid_vit1} and \cite{isaid_vit2}, however, argue that using attention-based CNNs are more effective towards the segmentation task. The authors of \cite{isaid_vit_imp} used ViT in their approach and subsequently got higher results than \cite{isaid_unet3} on the IoU score for the iSAID dataset: 67.20. They too used the UperNet as their training method, with the RingMo\cite{ringmo} as the train and the Swin-B ViT \cite{liu2021swin} as their backbone instead. This is impressive, but their model has 100 million parameters. We further prove that half of this is enough with our combined loss function to gain higher results on the dataset.

Xie et al. introduced SegFormer \cite{segformer} that utilize a similar technique to MaskFormer \cite{segformer} for semantic segmentation tasks. Their model performed exceptionally well with high efficiency in terms of performance to efficiency ratio but don't provide findings in terms of the iSAID datasets in their paper.

Hanyu et al. \cite{isaid_vit4} proposed a 3 ViT-based models: AerialFormer-T, AerialFormer-S, and AerialFormer-B with 42.7M, 64.0M, and 113.8M trainable parameters, respectively. Their models achieve the mIoU score of 67.5, 68.4, and 69.3 in the given three models. This score is higher than the one reported by \cite{isaid_vit_imp}\cite{isaid_unet3}. Even though the mIoU scores are highly desirable, the number of parameters in the AerialFormer-S and the AerialFormer-T are large with GLFLOPs of 49.0, 72.2, and 126.8, respectively, for the tiny, small, and base model. With respect to \cite{isaid_vit_imp} though, this is an incredible feat as the AerialFormer-T with 42.7M parameters outperformed the one in \cite{isaid_vit_imp} with 100M parameters. The authors don't mention the Dice scores for direct comparison, and we propose that our proposed model with 42M parameters can outperform the 113.8M parameters model with the help of our combined loss function. We also show that by using MaskFormer ViT as well, we can surpass the larger model.

Liu et al. \cite{wnet} proposed a CNN method for image segmentation on remote sensing images by dual path semantics approach. The authors devised a technique to make a new dual-path network structure, W-net, for the iSAID dataset and conducted an experiment to verify its generalization capacity. They conducted multiple ablation studies and deduced the highest amount of mIoU value of 63.68. Although respectable, we propose our model to yield better results than the one reported by the authors. The authors conclude that the size of the model impacts its inference time but do not mention the number of trainable parameters or the FLOPS in their model, which could be useful for common ground in model comparison.

From the literary analysis, past research has focused on improving the mIoU of their models (either ViT or CNN) by deploying new architectures, larger models, new loss functions or a combination of either of them. We find research gap on focusing on using a novel loss function that optimizes a relatively smaller CNN UNet based model against fine tuning a relatively larger MaskFormer ViT. We also show that by using our proposed weighted and combined loss function much smaller models can generalize and differentiate on background pixels and yield significantly closer results to models that are much larger than itself. This preserves the model's ability to retain high mIoU scores while not failing to classify pixels as background at the same time.

\section{Methodology}\label{methodology}
Fig. \ref{fig:methodology} summarizes the training and validation lifecycle of our proposed methodology. The red line segments show the work flow for the training flowchart of the UNet CNN model while the blue line segments show the training/finetuning flowchart for the MaskFormer ViT. As shown in Fig. \ref{fig:methodology}, the training lifecycle for our experiment includes five key steps: A. Dataset Information, B. Data Augmentation, C. Models, D. Loss Function, E. Validation Metrics, and F. Hyperparameters and Training Settings, which are all described below as their own subsections.

\subsection{Dataset Information}\label{dataset_info}
The iSAID dataset \cite{waqas2019isaid} is a benchmark in the remote sensing community due to its complex nature. The dataset is made of 2806 images from the DOTA dataset \cite{dota}. Out of these, 1411 images are training images, 458 validation images, and 937 unlabelled testing images. The resolution of the images range from $800 \times 800$ pixels to $4000 \times 13000$ pixels with 15 foreground and one background category. We only consider the foreground category while calculating our validation metrics scores.

The iSAID dataset is one of the most difficult dataset to generalize a model upon due to its complexities on the ranges of image size  and the small number of images. The authors from the dataset \cite{waqas2019isaid} acknowledge that any instance segmentation model that performs well on iSAID must have learned robust representations and effective localization capabilities. This would entail that a methodology that performs well on iSAID has a robust foundation to perform well on other segmentation tasks as well.

\subsection{Data Augmentation}\label{data_augmentation}
The data augmentation process followed the procedure shown in Algorithm~\ref{alg:augmentation}. The algorithm first takes anywhere between 6-28\% of the image and resizes it to the size of (512, 512) pixels. It then randomly flips the horizontal and vertical axis with a 50\% probability and rotates the image anywhere between (0 - 360)$^{\circ}$. Random brightness or contrast is added to the image and is normalized with the mean and standard deviation from the ImageNet dataset. The first step in the process ensures with maximum probability that each of the images we generate from any given image would be either scaled appropriately, have unique features, or both. After the combination of each of the given augmentations, each image can generate hundreds of new images, all of which would have some unique aspects to them.

\begin{algorithm}
    \caption{Data Augmentation Using Albumentations}\label{alg:augmentation}
    \begin{algorithmic}[1]
        \Procedure{Data Augmentation}{$Input Image$}
        \State $I \gets Input Image$
        \State $I \gets RandomResizeCrop(args, I)$
        \State $I \gets RandomVerticalFlip(args, I)$
        \State $I \gets RandomHorizontalFlip(args, I)$
        \State $I \gets RandomRotation(args, I)$
        \State $I \gets RandomBrightness(args, I)$
        \State $I \gets Normalization(args, I)$
        \State \textbf{return} $I$ 
        \EndProcedure
    \end{algorithmic}
\end{algorithm}

\subsection{Models}
We trained two models: the first is the UNet based on CNN, and the second is the fine-tuned MaskFormer ViT. For both the models, we use the gradient clipping algorithm by Pascanu et al.\cite{grad_clip} to account for exploding gradients. Through the trial and error approach, we concluded the maximum value of $threshold$ in algorithm \ref{alg:grad_clip} could be set to 3.0.
\begin{algorithm}[t]
    \caption{Gradient Clipping\cite{grad_clip}}\label{alg:grad_clip}
    \begin{algorithmic}[1]
        \State $\hat{g} \gets \frac{\partial L}{\partial \theta}$ \Comment{$\theta$ are the learnable parameters} 
        \If{$\|\hat{g}\| > threshold$}
            \State $\hat{g} \gets \frac{threshold}{\|\hat{g}\|} \cdot \hat{g}$
        \EndIf
    \end{algorithmic}
\end{algorithm}

Each model is different in its fundamental properties and is described in its own subsection below.

\begin{figure*}[!ht]
    \centering
    \includegraphics[width=0.9\textwidth, height=0.5\textheight]{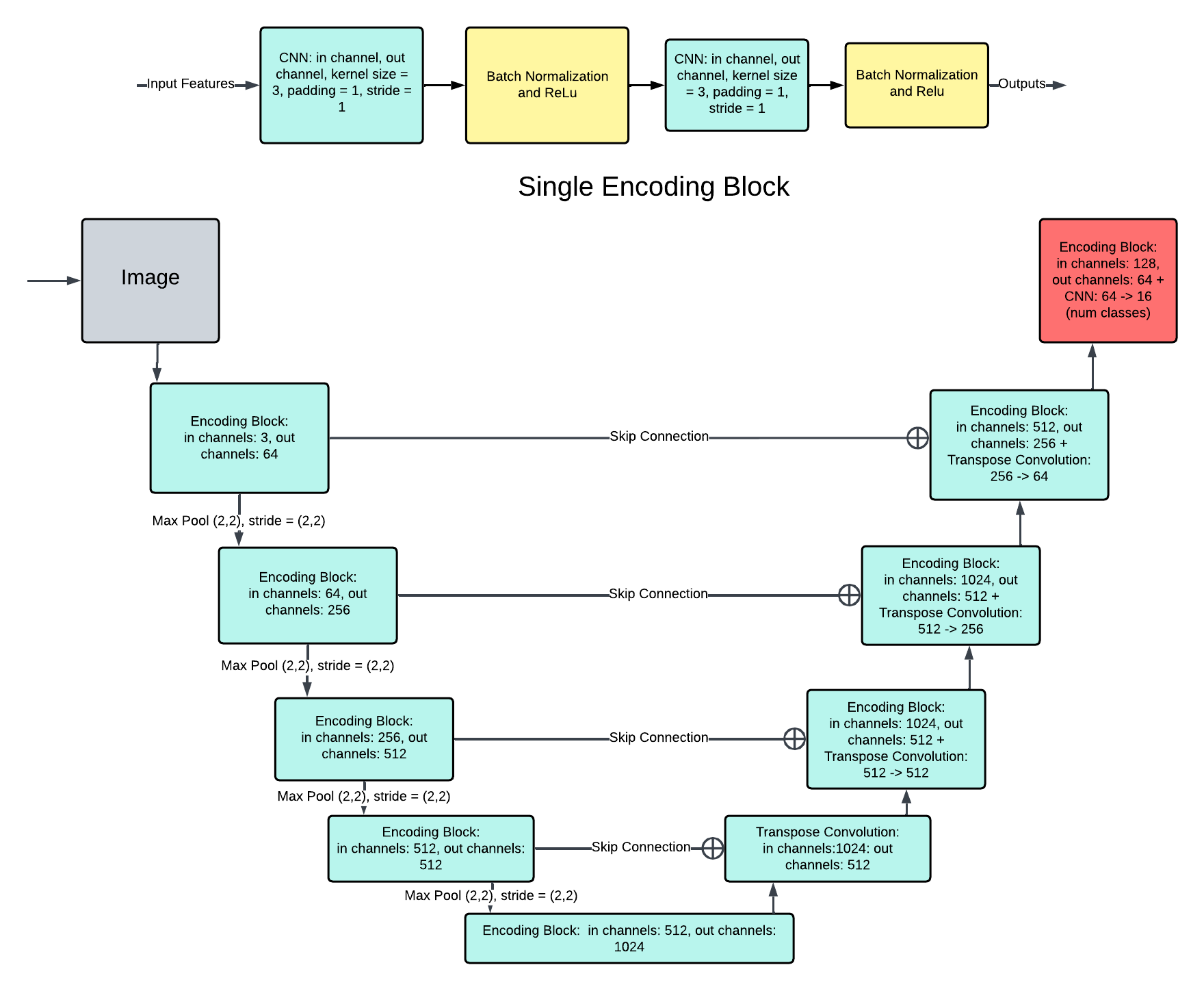}
    \captionsetup{justification=centering}
    \caption{Custom UNet CNN architecture}
    \label{fig:unet}
\end{figure*}
\subsubsection{UNet CNN}
Our version of the UNet architecture is a simple adaptation of a generic CNN-based UNet model with four skip connections. The overall architecture of the UNet CNN model is described in Fig. \ref{fig:unet}. We describe an encoding block and use it throughout the model to do all the feature extraction in both the convolution and transpose convolution layers. The total number of trainable parameters of Fig.\ref{fig:unet} amounts to 42.9M, with most of the learnable parameters concentrated in the bottleneck layers. We used gradient accumulation to mimic a batch size of 128 images and employed the mixed precision technique to quantize the forward operations to 16-bit floats. These two techniques make the forward process more efficient by skipping backpropagation for n accumulation steps and calculating the predicted mask with just 16-bit precision.

\begin{figure*}[t]
    \centering
    \includegraphics[width=\textwidth]{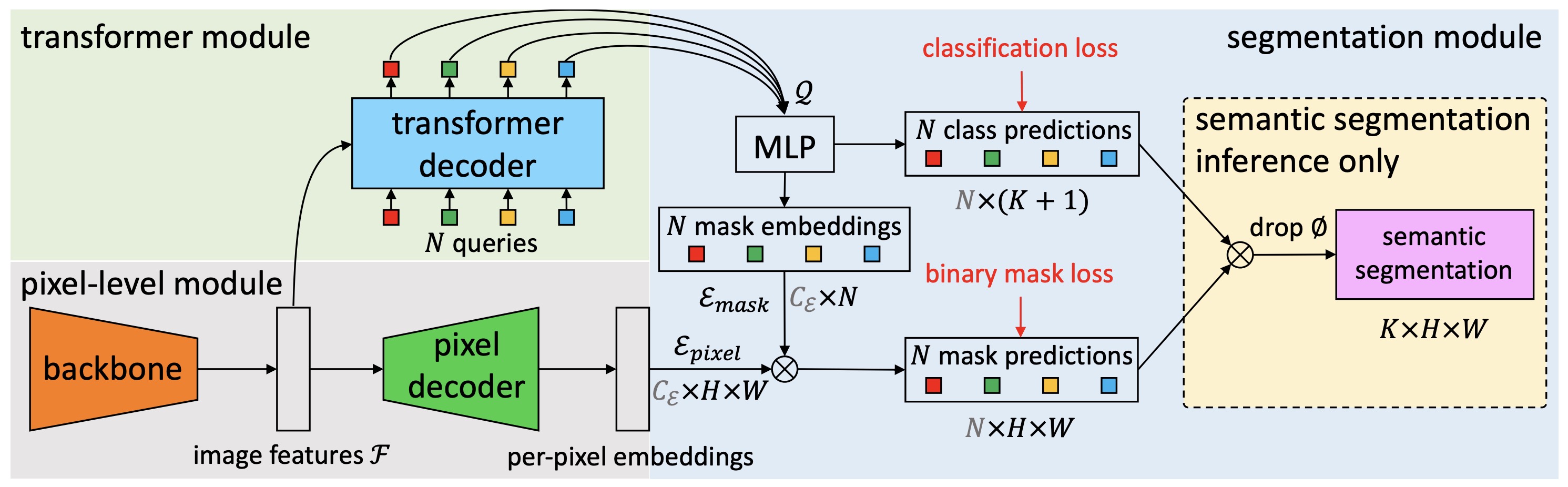}
    \captionsetup{justification=centering}
    \caption{MaskFormer ViT architecture from \cite{maskformer}}
    \label{fig:maskformer}
\end{figure*}

\subsubsection{MaskFormer ViT}

MaskFormer\cite{maskformer} is an algorithm that infuses pixel decoder and transformer decoder with a pre-trained backbone, usually a ResNet\cite{resnet} block. The image is forwarded into the backbone and the resultant is passed on to the the pixel decoder and transformer decoder. The pixel decoder would produce per-pixel embeddings, and the transformer's output after being sent to an MLP gives two results: the N class classification and the N mask embeddings. The N mask embeddings are then combined with the per-pixel embeddings, and the output is combined with the N class classification to get the final segmentation. This is represented by Fig \ref{fig:maskformer} extracted from the original paper \cite{maskformer}. 

The number of parameters for MaskFormer depends upon the backbone used and ranges from 41M to 212M parameters. For our experiment, we used Facebook's Swin Large architecture, pre-trained on imagenet22k, and fine-tune it for the 15 classes in the iSAID dataset, resulting to approximately 200M parameters. Since most of the parameters are frozen and only the top layer trained, the training efficiency was faster than the UNet counterpart.

\subsection{Loss Function}

Authors form \cite{maskformer} employ a novel loss function, which is a combination of the Dice loss function and focal loss function \cite{focalloss}. Therefore, we used the same implementation for the fine tuning process. For the UNet, we come up with our proposed weighted loss function which focuses on maximizing mIoU, Dice and minimizing cross-entropy.

Take an image mask $A$ and probability distribution, or the predicted probability, of the mask $B$ such that $A,B$ $\in \mathbb{R}^{C*H*W}$ where $C$, $H$ and $W$ are the channel width and height of the mask respectively. Then in order to maximize the $mIoU$ we minimize $1-IoU$ as our loss function \cite{iouloss}.

\begin{equation}
   L_{iou} = 1 - \frac{A\cap B}{A \cup B} 
\end{equation}

In order to make the loss function differentiable, we replace the nondifferentiable operations of the bitwise intersection (AND) operation and the union (OR) operation with multiplication and addition operations.
\begin{equation}\label{iou}
   L_{iou} = 1 - \frac{A * B}{A + B - (A*B)}
\end{equation}

Similarly, in order to maximize the Dice score, we need to minimize $1-Dice$ score. Since the bitwise intersection function is not differentiable, we replace the intersection with equivalent multiplication, so the loss function $L_{dice}$ becomes the following \cite{diceloss}:

\begin{equation}\label{dice}
   L_{dice} = 1 - \frac{2 * A * B}{|A| + |B|}
\end{equation}

We also use a weighted cross-entropy loss function to maintain the entropy of our predictions, so the third part of our function becomes $L_{ce}$ \cite{crossentropy}.

\begin{table*}[!ht]
\caption{IoU Scores Comparison by Category }
\label{tab:performance}
\centering
\scriptsize 
\renewcommand{\arraystretch}{1.3}
\begin{tabularx}{\textwidth}{c c c c | c c c c c c c c c c c c c c c}
\toprule
\multirow{2}{*}{\textbf{Method$\dagger$}} & \multirow{2}{*}{\textbf{\#params}} & \multirow{2}{*}{\textbf{Year}} & \multirow{2}{*}{\textbf{mIoU}} & \multicolumn{15}{c}{\textbf{IoU per Category in \%}} \\ \cline{5-19}
 &  &  &  & \textbf{PL} & \textbf{BD} & \textbf{BR} & \textbf{GTF} & \textbf{SV} & \textbf{LV} & \textbf{SH} & \textbf{TC} & \textbf{BC} & \textbf{ST} & \textbf{SBF} & \textbf{RA} & \textbf{HA} & \textbf{SP} & \textbf{HC} \\ \hline
Plain ViT \cite{isaid_vit_imp} & 100M & 2022 & 67.20 & \multicolumn{15}{c}{\textbf{Not Reported}} \\

AANet \cite{aanet} & 29.2M & 2022 & 66.6 & 84.6 & 80.5 & 40.2 & 60.5 & 48.7 & 63.2 & 71.2 & 88.8 & 65.4 & 65.7 & 73.5 & 72.4 & 57.2 & 52.3 & 41.8 \\

AF-T \cite{isaid_vit4} & 42.7M & 2023 & 67.5 & 86.1 & 77.5 & 45.3 & 57.5 & 52.6 & 67.0 & 68.6 & 88.8 & 63.4 & \textbf{74.9} & \textbf{75.1} & 73.0 & 58.2 & 50.5 & 42.0 \\

AF-S \cite{isaid_vit4} & 64M & 2023 & 68.4 & 86.5 & 78.8 & 44.8 & 59.5 & 53.6 & 66.5 & 72.1 & \textbf{89.2} & 66.5 & 74.1 & 77.0 & 74.0 & 60.9 & 52.1 & 40.0 \\

AF-B \cite{isaid_vit4} & 113.8M & 2023 & 69.3 & 86.5 & \textbf{81.5} & 46.8 & \textbf{65.0} & 53.7 & 67.8 & \textbf{75.1} & \textbf{89.8} & 62.4 & 76.3 & 78.3 & 66.1 & 60.8 & 52.4 & 46.7 \\

Ringmo \cite{ringmo} & 87.6M & 2023 & 67.2 & 85.7 & 77.0 & 43.2 & 63.0 & 51.2 & 63.9 & 73.5 & 89.1 & 62.5 & 73.0 & 78.5 & 67.3 & 58.9 & 48.9 & 40.1 \\

SAMRS-V \cite{isaid_unet3} & - & 2023 & 66.0 & 86.0 & 79.3 & 42.5 & 65.3 & 53.2 & \textbf{68.4} & 75.9 & 89.5 & 63.7 & 75.4 & 79.2 & 69.8 & 59.5 & 49.4 & 37.5 \\

SAMRS-C \cite{isaid_unet3} & - & 2023 & 62.54 & 82.9 & 76 & 40 & 61.9 & 48.5 & 63.8 & 69.8 & 87.7 & 58.1 & 71.1 & 76.1 & 70.2 & 56 & 48.6 & 27 \\

W-Net-C \cite{wnet} & - & 2023 & 56.7 & 50.6 & 59.7 & 61.5 & 43.2 & \textbf{64.7} & 55.9 & \textbf{88.9} & 72.1 & 44.4 & 42.1 & 79.9 & 56.6 & 67.7 & 29.5 & 18.6 \\ \hline

\textbf{Ours (CNN)} & 42.9M & 2024 & \textbf{73.4} & 64.6 & \textbf{89.7} & \textbf{91.7} & \textbf{84.8} & 43.9 & 52.5 & 63.8 & 71.8 & 71.2 & \textbf{82.2} & \textbf{88.0} & \textbf{94.1} & 64.5 & \textbf{89.0} & 49.2 \\

\textbf{Ours (ViT)} & 200M & 2024 & \textbf{82.48} & \textbf{92.8} & 77.9 & 82.3 & 79.9 & 53.8 & 62.7 & \textbf{82.4} & \textbf{90.2} & \textbf{83.8} & 81.6 & 79.7 & \textbf{93.3} & \textbf{88.6} & \textbf{93.1} & \textbf{95.3} \\

\bottomrule
\end{tabularx}
\begin{tablenotes}
\item[a] Abbreviations: PL = Plane, BD = Baseball Diamond, BR = Bridge, GTF = Ground Track Field, SV = Small Vehicle, LV = Large Vehicle, SH = Ship, TC = Tennis Court, BC = Basketball Court, ST = Storage Tank, SBF = Soccer Field, RA = Roundabout, HA = Harbor, SP = Swimming Pool, HC = Helicopter. $\dagger$: V=ViT, C=CNN, T=Tiny, S=Small, B=Base.
\end{tablenotes}
\end{table*}

\begin{figure*}[!ht]
    \centering
    \includegraphics[width=0.7\textwidth]{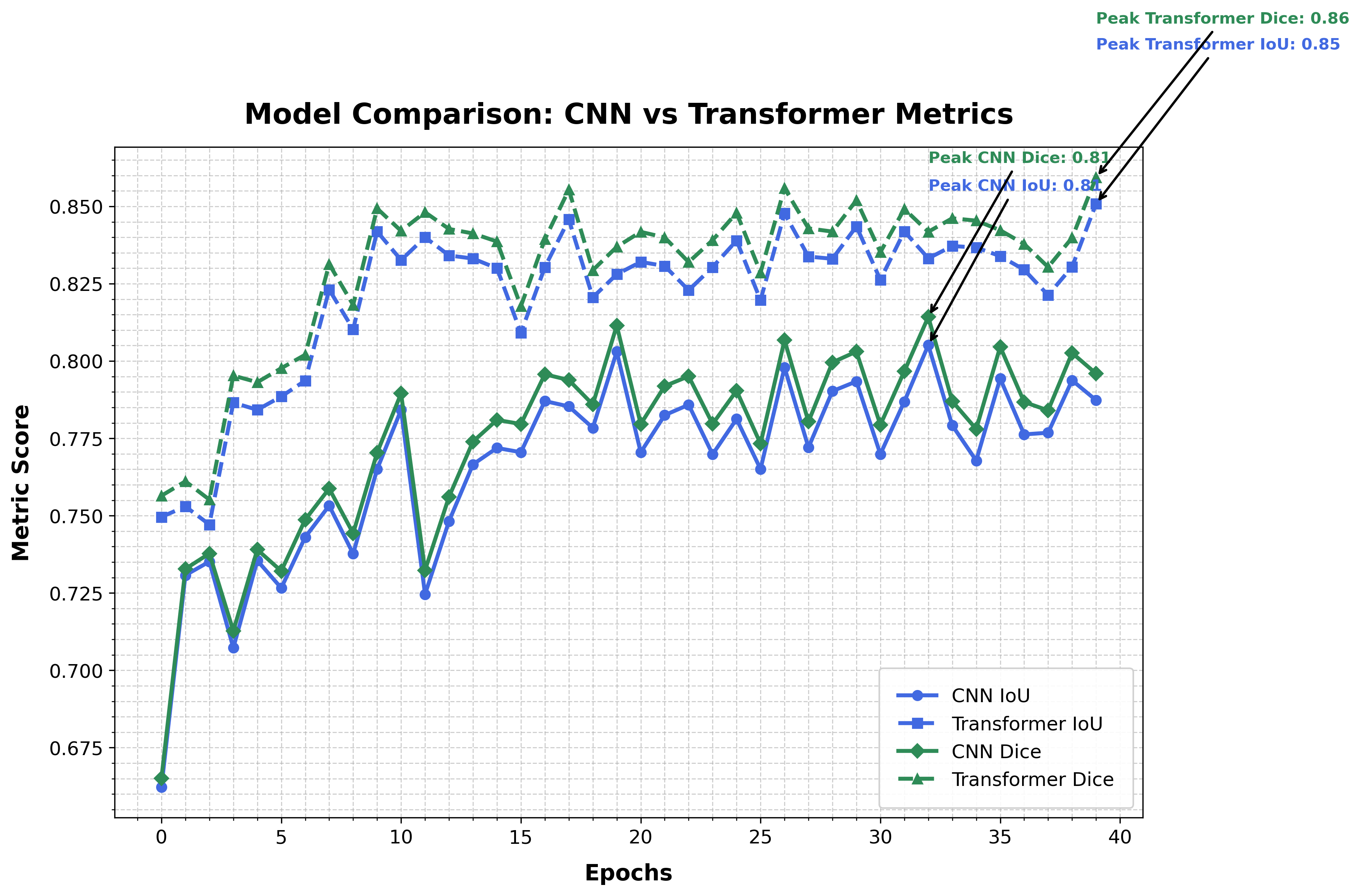}
    \caption{Comparison of Metrics over Epochs}
    \label{fig:metrics}
\end{figure*}

\begin{equation} \label{crossentropy}
L_{ce} =   \beta *A \log(B) + (1 - \beta)(1 - A) \log(1 - B) 
\end{equation}
where \( \beta \) is the weight hyperparameter (0.15 for unlabelled class and 1 for the rest),

Combining the loss functions (\ref{iou}) (\ref{dice}) (\ref{crossentropy}) together with weights $\lambda_{iou},$ $\lambda_{dice},$ $ \lambda_{ce}$ respectively we get the total combined loss $L$.
\begin{equation}\label{combined}
    L = \lambda_{iou} * L_{iou} + \lambda_{dice} * L_{dice} + \lambda_{ce} * L_{ce} 
\end{equation}
where \( \lambda_{iou} \) = 0.8, \( \lambda_{dice} \) = 1 and \( \lambda_{ce} \) = 10 were selected by trial and error experimentation.

We also did not calculate the loss functions $L_{iou}$ and $L_{dice}$ for the unlabeled class, unlike with $L_{ce}$ because during inference we need not care about properly labeling the unlabeled class, but the model still needs some hint of guidance of the types of patterns it needs to avoid during the training phase. 

We choose the minimal value of 0.15 for \(\beta\) in $L_{ce}$. This loss function in Eq. \ref{combined} represents the final form of our proposed loss.

\subsection{Validation Metrics}
We validate the results produced by our model for $C$ classes using the mIoU score and Dice score described as follows \cite{iouloss}\cite{diceloss}:

\begin{equation}\label{mIoU}
   mIoU = \frac{1}{C}\sum_{i=0} ^{i=C} \frac{A_i\cap B_i}{A_i \cup B_i} 
\end{equation}
\begin{equation}\label{Dice}
   Dice = \frac{1}{C}\sum_{i=0} ^{i=C} \frac{2*(A_i\cap B_i)}{|A_i| + |B_i|}
\end{equation}

\subsection{Hyperparameters and Training Settings}
Adam\cite{adam} was chosen as the optimizer for both models because of its ability to fine-tune the learning rate over time. We trained the models for 40 epochs, each with an initial learning rate of $10^{-3}$ on a Nvidia A40-48Q GPU with 48 GB VRAM. The initial learning rate was chosen so it could easily increase or decrease its value depending on the model's performance. As Adam updates the learning rate depending on the gradients automatically, all we had to ensure was that the loss was decreasing and the validation metrics were increasing. The models were then  validated at the end of each epoch under the Dice score and IoU scores described in (\ref{mIoU})(\ref{Dice}).

\begin{table}[t]
\caption{Models Inference Efficiency Test Results}
\label{tab:Training time}
\centering
\renewcommand{\arraystretch}{1.1}
\begin{tabularx}{\columnwidth}{c c c c}
\toprule
\textbf{Model Name} & \textbf{\# of Parameters} & \textbf{FLOPS} & \textbf{Inference Time*}\\
\hline
UNet CNN & 42.9M & 460.10 G & 0.19s\\
MaskFormer & 200M & 232.40 G & 0.29s
\end{tabularx}
\begin{tablenotes}
\item[a] * Inference Time calculated on 6 images
\end{tablenotes}
\end{table}

\section{Results}\label{results}

\begin{figure*}[!ht]
    \centering
    \includegraphics[width=\textwidth, height=0.8\textheight]{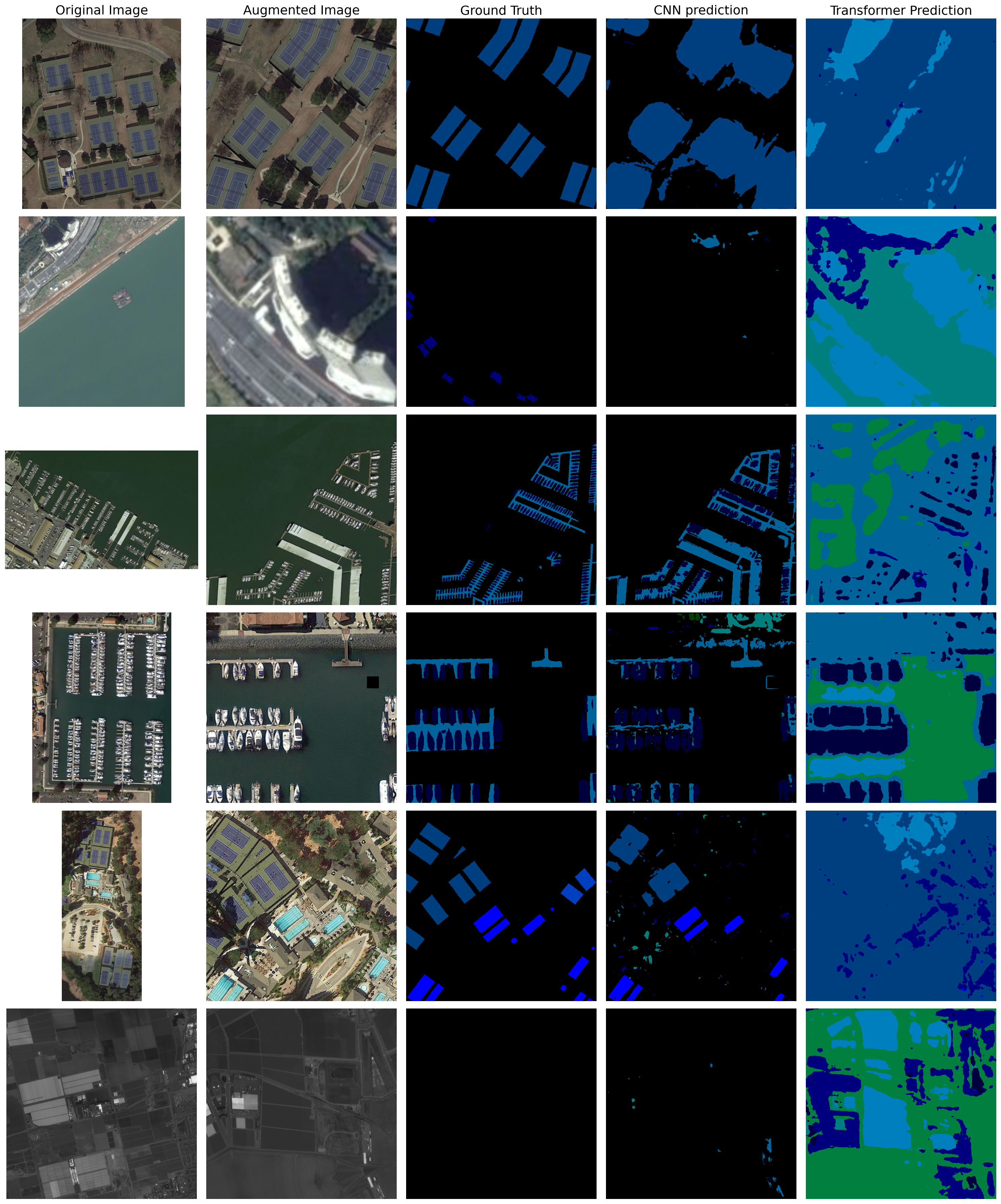}
    \caption{Sample Visualization of model's output}
    \label{fig:sample}
\end{figure*}

\begin{figure*}[!ht]
    \centering
    \includegraphics[width=0.95\textwidth]{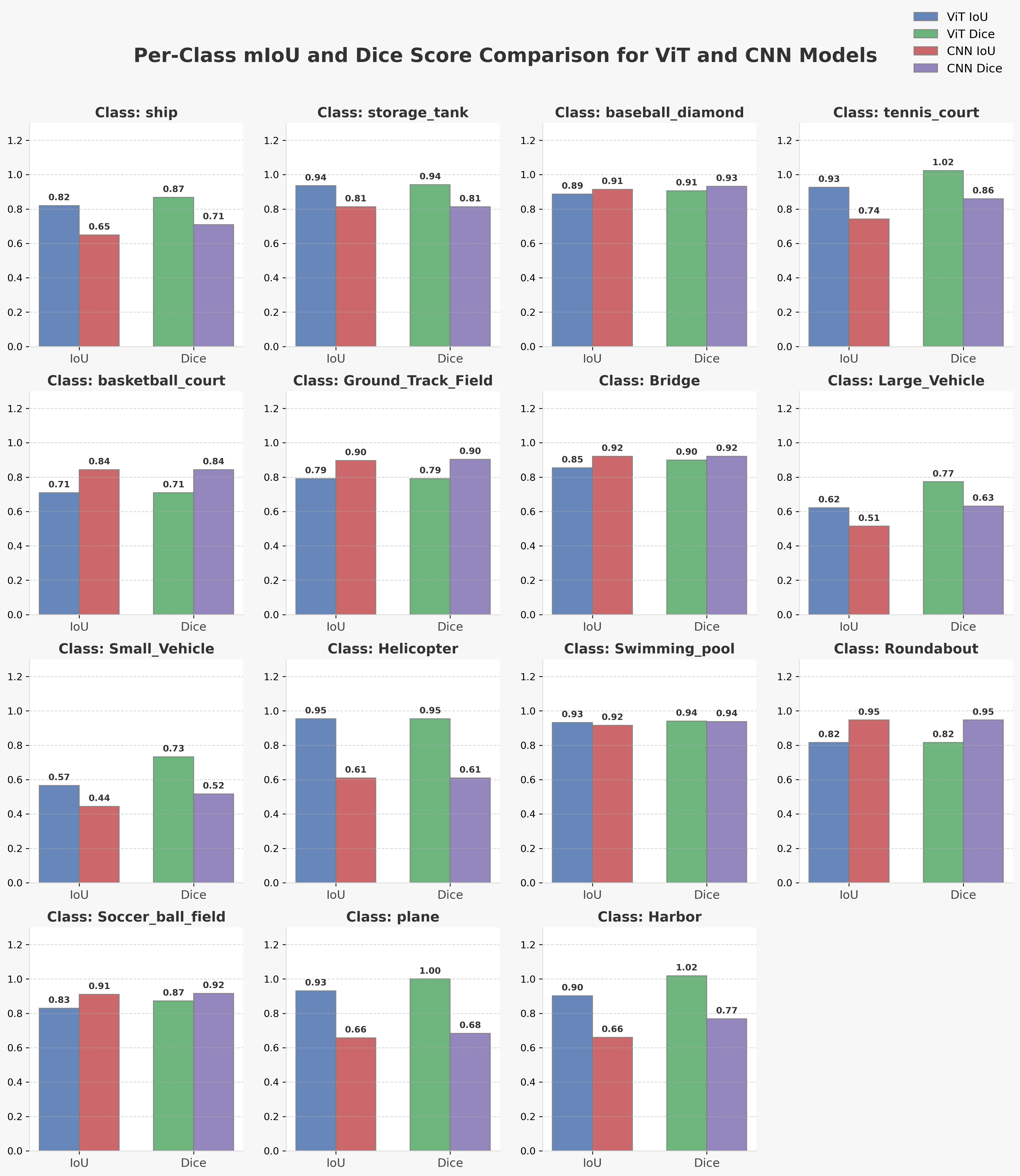}
    \caption{Class-wise comparison of IoU and Dice Scores}
    \label{fig:per_class_dice}
\end{figure*}

Our findings based on the methodology graph shown in Fig \ref{fig:methodology} are presented in this section. 

First, we list out the two model's efficiency information in Table \ref{tab:Training time}. During training, most of the time was consumed by the data augmentation technique described in \ref{data_augmentation}, which took 65 seconds per batch for a batch size of 128. During inference on six images, however, we can see that having more numbers of parameters hurts the efficiency of the MaskFormer model. This moves us into the question of FLOPS vs Inference Time comparison during inference. From Table\ref{tab:Training time}, we notice that UNet CNN has a higher FLOPS and a lower inference time (460.10 G vs 0.19s) as compared to that of MaskFormer (232.40G vs 0.29s). This indicates that on average the UNet model performs better in terms of efficiency by making full utilization of the resources for shorter bursts of time. For where segmentation tasks are mostly used like in UAV, cars, UGV and similar tools, having models which can give output as soon as possible with relatively good accuracy is more important than waiting a few extra seconds to gain a 3-5\% increase in the output. This indicates that the UNet CNN model could be a better alternative just by looking at the inference table.

The MaskFormer ViT model, however, churned better results in terms of mIoU and Dice scores Fig \ref{fig:metrics}. The Dice Score and mIoU for the MaskFormer soared from mid 70\% to the range of 80\% by the end of the last epoch, whereas for UNet CNN, it started low on around 65\% and reached the peak of 81\% and stablized at around 78\% during training over all classes Fig. \ref{fig:metrics}. The Dice score and mIoU scores have a strong correlation between them, so we further compare the mIoU scores of the model against our references and recent research works. 

In Table \ref{tab:performance}, recently published and reference works in the topic were selected whose models either have a comparable number of parameters, architectures, or both as ours to be compared with respect to the mIoU scores and per category IoU scores. We have bolded the previous research with comparable or better results in terms of IoU per category in the table. The key advantage here is that the background class is not taken into consideration in any, ours or the previous, of the evaluation. This gives a good advantage to all the models but doesn't truly show the strong capacity of the weighted combined loss function to generalize and bring about meaningful maps which we further discuss next. Both the models performed exceptionally well, especially in terms of mIoU, but do have some pitfalls in terms of IoU per category where generalization to some classes like small vehicle and large vehicle were better comprehended by previous works. The only problem with this evaluation is we don't know the exact number of parameters, the inference time and the FLOPS for the models which did better in these specific category in Table \ref{tab:performance}. 

It can be noted that the mIoU and Dice scores for the MaskFormer ViT are within the 10\% upper range of the UNet CNN model, even while having ~5 times more parameters in it. From further analysis of the Fig \ref{fig:sample}, however, we can notice that even though the Maskformer has higher metrics as compared to the UNet CNN, it has failed to rationalize the background class due to its low tolerance or importance towards the background or none class objects in the image. This entails that without a pixel mask that tells the model on which pixels to predict and which to not predict, the results are not feasible on the model unless trained without any background class like in datasets where every pixel is classified into a certain group. This also stems from the fact that we calculate the scores on only the valid mask of pixels, i.e., the pixels that belong to the background class are ignored during computation of mIoU and Dice as discussed earlier in section \ref{methodology}.

Fig \ref{fig:per_class_dice} represents the per class Dice and IoU scores for the two proposed models. The correlation between the two scores shown in Fig \ref{fig:metrics} is further solidified by the class-wise comparison. The worst performance, according to Fig \ref{fig:per_class_dice}, is on the small vehicle class, and both models generally perform well in classes that include general landmarks like baseball diamonds, tennis courts, basketball courts, and ground track fields. Other easily recognizable items, which rarely change their shape, form, and size, like a bridge and roundabout, are also among the best-predicted classes by both models.

Our UNet CNN model's metrics on the validation set surpassed the performance of similarly comparable references on the test set, and the MaskFormer, with its greater number of parameters, surpassed the UNet CNN model as well. It can be noted, however, that the mIoU per category of our, or the reference works, were not uniformly distributed, and there's a high affinity towards objects that tend to be larger in size and more frequent within the dataset. Fig \ref{fig:sample} shows sample prediction of randomly selected data against the UNet CNN model and the MaskFormer ViT. The augmentation process shows a high yield of variability in the images, which is capable of generating multiple almost unrecognizable images from a single one. The ground truth was taken from the validation dataset itself, and the UNet and Maskformer's predictions followed to the right. The segmentation maps in Fig \ref{fig:sample}, based on the category, show its consistency with the mIoU presented in Table \ref{tab:performance} and the average overall metrics shown in Fig \ref{fig:metrics}. This correlation fgvrther validates the experimental design of the combined loss and puts UNet CNN directly against the MaskFormer ViT yielding not only good results in terms of numbers but also in terms of generalization on unseen images, the background class and high fidelity of the segmentation maps.

\section{Conclusion and Discussions}\label{conclusion}

We successfully show that the introduction of the combined weighted loss function helps the model to make stronger predictions and yield better results. One caveat of the findings could be that we need to find a stronger way to de-segment the testing images into multiple patches of smaller images, which could then be fed into the model and then rearranged later to complete the original images for the testing set, as we learn that simply rescaling does affect the model negatively; in other words, the image loses key patterns and features with simple rescaling in data augmentation. We show the number of parameters required to make a robust remote imagery sensing segmentation model does not need to be over 50M, given that it can directly influence the FLOPS and inference time. We have shown our results in context of previous literature and proved that the usage of the proposed loss does affect the performance model positively and with great significance. 

Although the larger training time could be attributed to the data augmentation process, future research could look into streamlining this entire pipeline as a whole. The next steps in the field would entail decoupling the image into multiple smaller fragments and recoupling into the original image to have a robust prediction in images of any shape and size. Nevertheless, we introduced a strong novel combined and weighted loss function to compare UNet CNN with transfer learning-based ViT and show their performance against similar state-of-the-art segmentation models. We show with the use of the new loss function, the CNN model yields comparable results on the metrics with better generalization capacity during inference on unseen data. We also show that the model was better able to make use of the hardware on full capacity yielding the mentioned results while demanding significantly less amount of time.

\bibliographystyle{IEEEtran}

\end{document}